\documentclass[conference]{IEEEtran}
\IEEEoverridecommandlockouts

\usepackage{amsmath,amssymb,amsfonts}
\usepackage{algorithmic}
\usepackage{graphicx}
\usepackage{textcomp}
\usepackage{xcolor}
\usepackage{hyperref}
\usepackage{dblfloatfix}

\usepackage{amsfonts}
\usepackage{booktabs}
\usepackage{siunitx}

\hypersetup{
    colorlinks=true,
    linkcolor=blue,   
    urlcolor=cyan,
    citecolor = black,
    }

\usepackage{biblatex}
\usepackage[export]{adjustbox}
\usepackage{graphicx} 
\usepackage[font=small,labelfont=bf]{caption} 

\def\BibTeX{{\rm B\kern-.05em{\sc i\kern-.025em b}\kern-.08em
    T\kern-.1667em\lower.7ex\hbox{E}\kern-.125emX}}
    
\begin{document}

\title{Leaf-Based Plant Disease Detection and Explainable AI}

\author{
\IEEEauthorblockN{Saurav Sagar$^{1}$, Mohammed Javed$^{*1}$, David S Doermann$^{2}$}

\IEEEauthorblockA{
\textsuperscript{1}Computer Vision and Biometrics Lab, Department of IT, Indian Institute of Information Technology, Allahabad, India \\
\textsuperscript{2}Department of CSE, University at Buffalo, Buffalo, NY, USA \\
\text{mit2021079@iiita.ac.in, javed@iiita.ac.in (*Corresponding author), doermann@buffalo.edu}}
}

\maketitle

\begin{abstract}
The agricultural sector plays an essential role in the
economic growth of a country. Specifically, in an Indian context,
it is the critical source of livelihood for millions of people living in rural areas. Plant Disease is one of the significant factors affecting the agricultural sector. Plants get infected with diseases for various reasons, including synthetic fertilizers, archaic practices, environmental conditions, etc., which impact the farm yield and subsequently hinder the economy. To address this issue, researchers have explored many applications based on AI and Machine Learning techniques to detect plant diseases. This research survey provides a comprehensive understanding of common plant leaf diseases, evaluates traditional and deep learning techniques for disease detection, and summarizes available datasets. It also explores Explainable AI (XAI) to enhance the interpretability of deep learning models' decisions for end-users. By consolidating this knowledge, the survey offers valuable insights to researchers, practitioners, and stakeholders in the agricultural sector, fostering the development of efficient and transparent solutions for combating plant diseases and promoting sustainable agricultural practices.\\
\end{abstract}

\section{\textbf{INTRODUCTION}}
The agriculture sector is a vital component of the economies of both developed and developing nations. In India, agriculture occupies a significant portion of the country's geographical area, as reported by the Department of Agriculture Farmers Welfare (2022-2023) in their annual report.
With India's abundant arable land and favorable agro-climatic zones, the agricultural industry plays a pivotal role in sustaining the livelihoods of a large rural population. However, the ever-increasing population poses a formidable challenge to the sector, as it necessitates a substantial increase in food production to meet the growing demand. India, the most populous country in the world with a population of approximately 1,380.04 million, is particularly affected by this challenge. According to the
report, the demand for food grains in India is projected to rise by 53 million tonnes by 2030. Enhancing the productivity of the agricultural sector becomes imperative to address this demand. However, several factors, including soil and topography, climate change, and improper use of fertilizers, can adversely affect crop productivity. Among these factors, disease attacks on plants emerge as a critical concern.

Disease attacks on plants can potentially cause severe damage, leading to abnormal functioning and reduced productivity. Leaves, the primary site of environmental interaction, are particularly susceptible to various diseases. Timely detection and accurate classification of leaf diseases are crucial for effective disease management, as they enable targeted interventions and timely treatment measures. Traditionally, disease detection relied on expert manual analysis, which is time-consuming and labor-intensive. Alternatively, chemical-based pesticides have been widely used but can harm plant quality and beneficial soil microorganisms. Recent advancements in computer vision and AI techniques offer promising solutions for automating the detection and classification of leaf diseases. These approaches leverage deep learning models, particularly Convolutional Neural Networks (CNNs) and Transformer-based architectures, to analyze leaf images and accurately identify disease patterns. These deep-learning methods have demonstrated impressive performance in numerous crops, including apples, tomatoes, corn, and grapes. However, these models often lack interpretability despite their effectiveness, making understanding the reasoning behind their decisions challenging.

In this survey paper, we plan to summarize the various commonly occurring leaf diseases that infect plants and the available datasets and state-of-the-art techniques for detecting infected leaf diseases. Furthermore, we intend to introduce Explainable AI (XAI) in plant leaf-based disease detection and classification. The goal is to enhance the transparency and interpretability of deep learning models by generating XAI-based solutions tailored explicitly for CNN and Transformer models. The study also underscores the motivation for using XAI in plant leaf disease detection and highlights possible future research directions.

The entire paper is organized into six sections. Section II provides a brief overview of different plant leaf diseases and the detection techniques utilizing recent technology. Sections III and IV discuss the fundamental principles of XAI in the context of leaf disease detection. Section V offers insights and pointers for future plant leaf disease detection research. Finally, Section VI concludes this survey.

\section{\textbf{OVERVIEW ON LEAF BASED PLANT DISEASE DETECTION}}

All living creatures, like plants, animals, and humans, are disease-prone. People worldwide working in the field of agricultural science and management are looking for a new advanced solution to tackle plant disease attacks, which are capable of causing lethal damage to agricultural productivity. Thus, several branches of science work together to control the spread of plant leaf disease attacks to ensure a sufficient food supply for the world's growing population. Plant disease can cause various symptoms that impair a plant's structure (like leaf, stem, root, etc.) and ability to yield, reproduce or grow normally. There are a variety of plant diseases whose occurrence varies from season to season according to the change in weather conditions in the presence of a specific type of pathogens.\newline
\\This section is divided into three parts to discuss the common leaf diseases and available datasets and highlight critical research contributions in leaf-based plant disease detection.

\subsection{\centering\normalfont{\textbf{Common Plant Leaf Diseases}}}
Plant diseases primarily affect leaves and roots, stems, and fruits. Leaf diseases are the most common type of plant disease. Fungicides, bactericides, and resistant varieties are typically used to control them. Some of the most common leaf diseases are mentioned below.

\begin{itemize}
\item\textbf{Scab:}
It is a fungal disease with several host-specific characteristics capable of infecting a single plant. It is a common disease found in apple trees. Leaves infected with scabs are initially olive green in color (Fig. 1.a), which turns yellowish before they fall.

\begin{figure} 
\small
\centering
\begin{tabular}{c c} 

\includegraphics[width=3.5 cm, height=4cm]{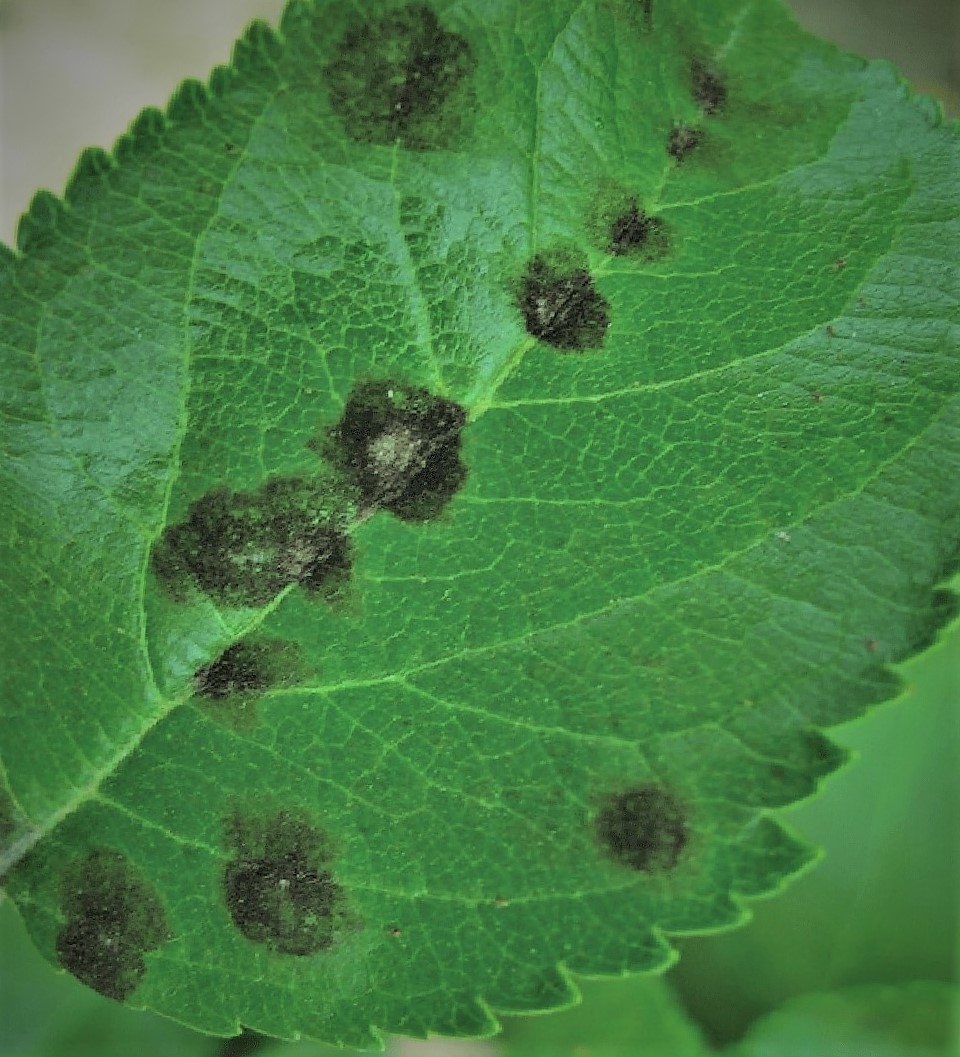} & 
\includegraphics[width=3.5 cm, height=4cm]{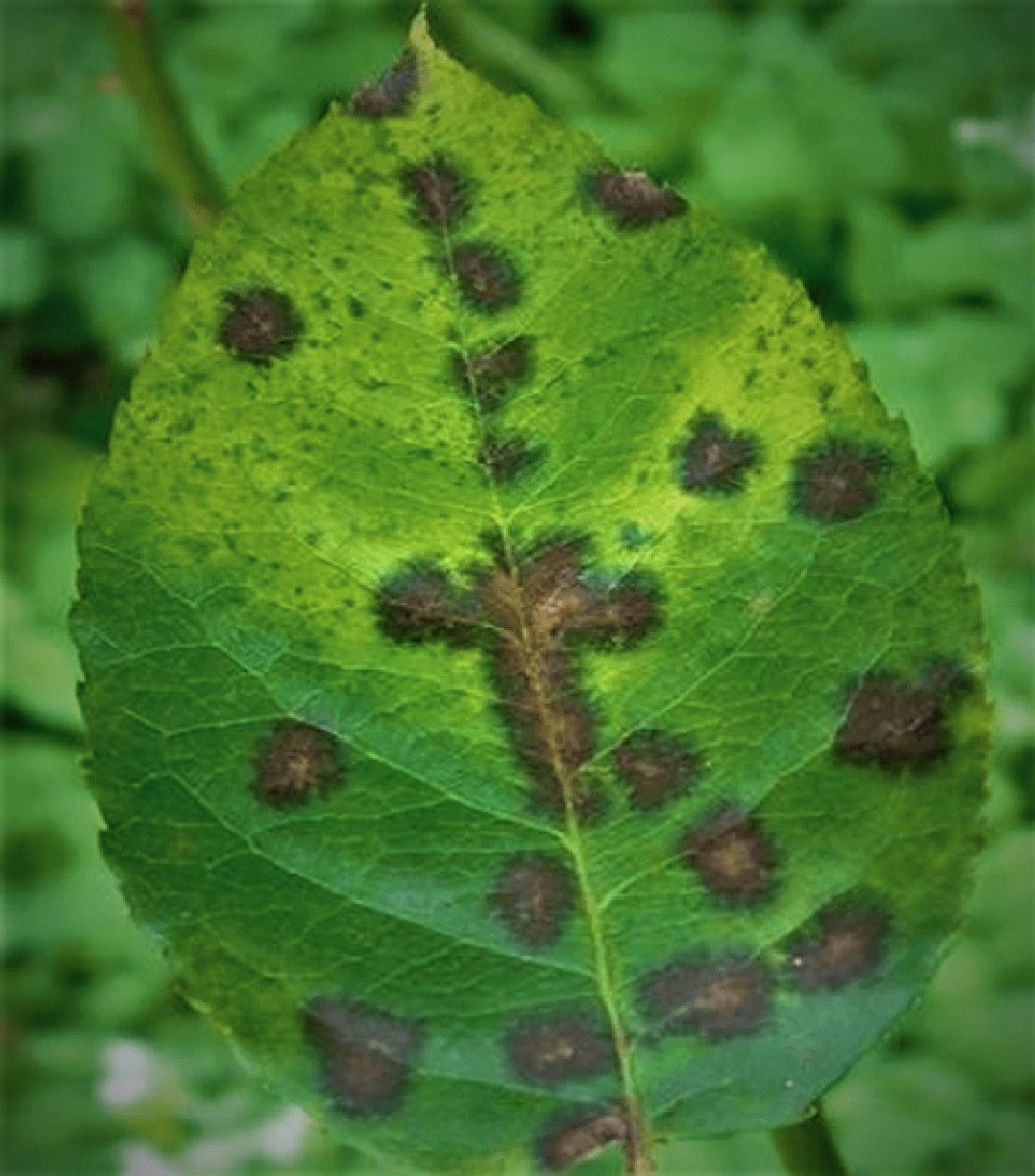}
\\

\text{a. Apple Scab}  & \text{b. Rose Black Spot} 
\\
\\

\includegraphics[width=3.5 cm, height=4cm]{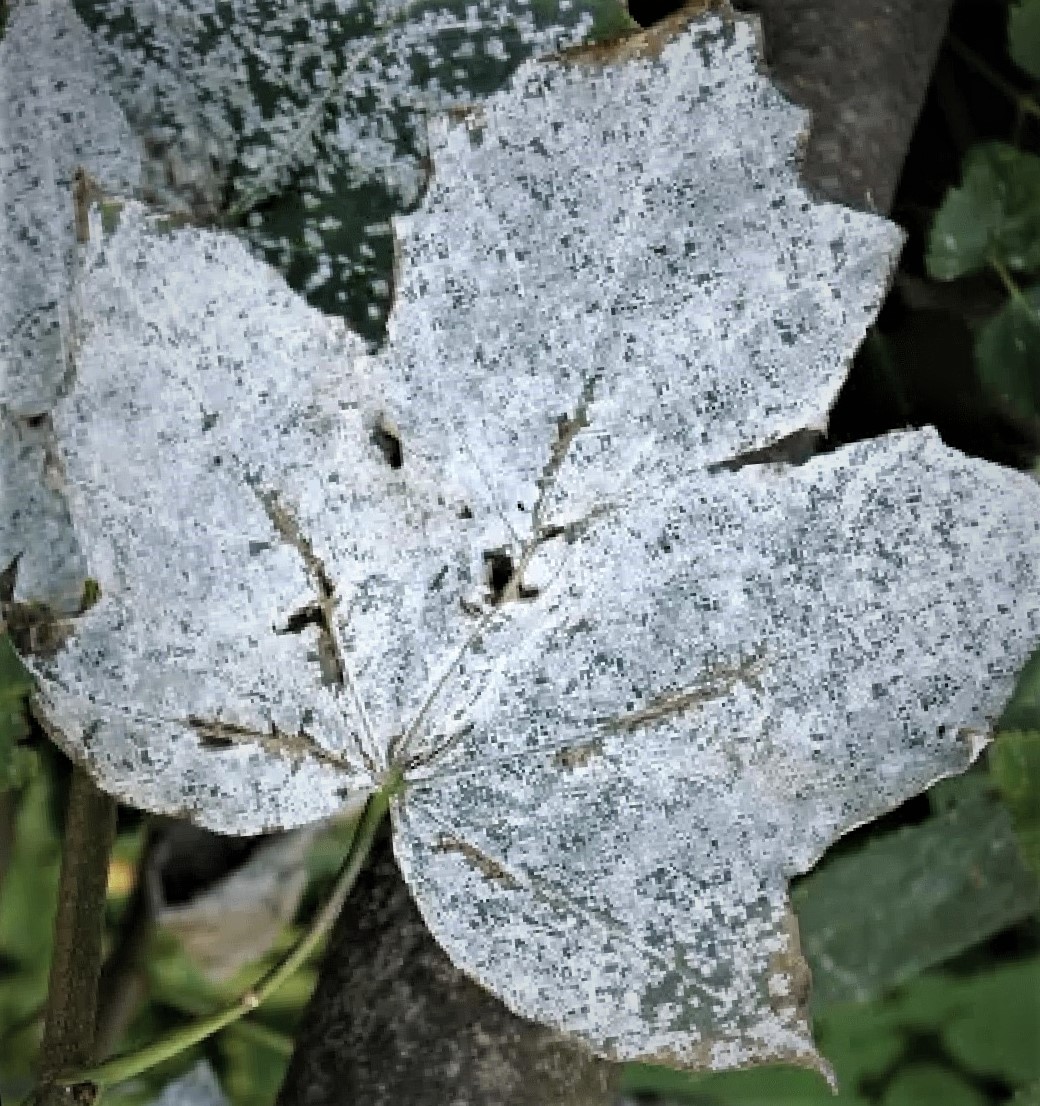} & \includegraphics[width=3.5 cm, height=4cm]{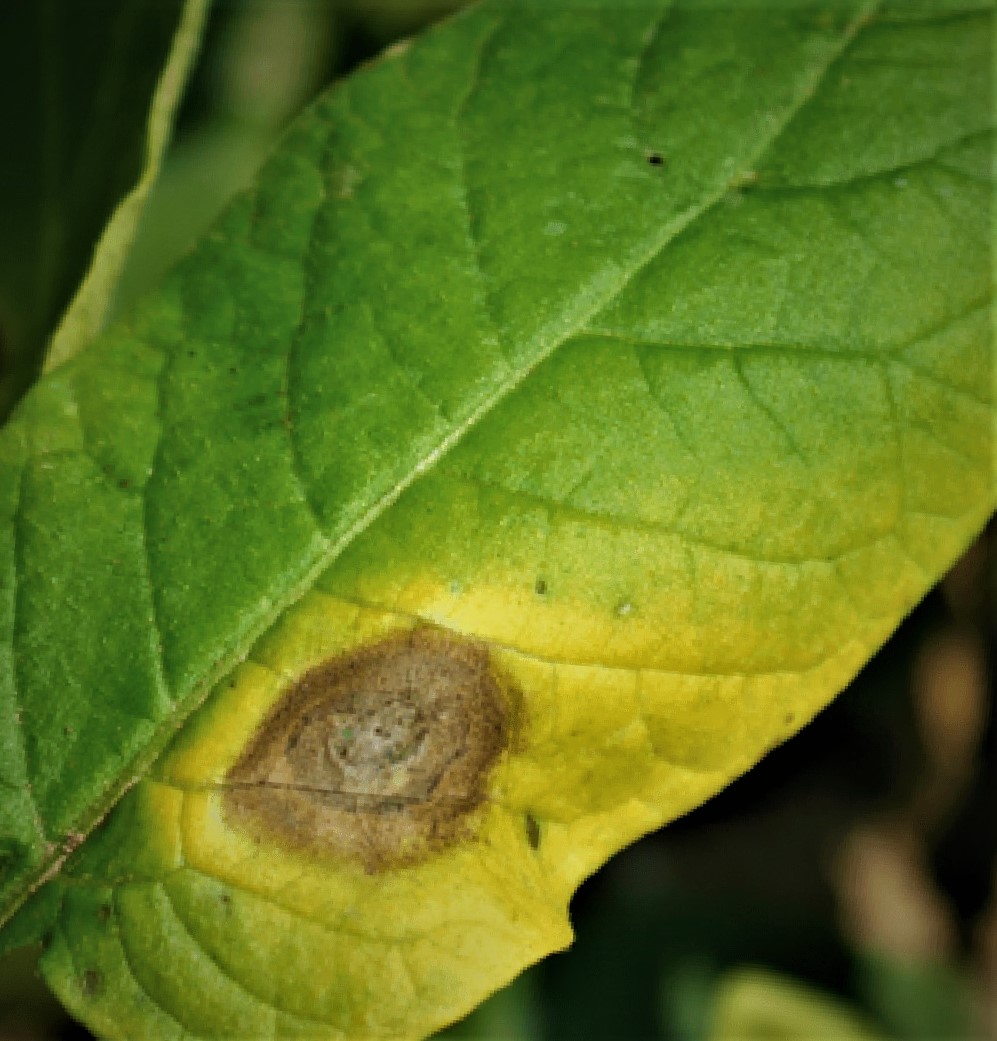}
\\

\text{c. Squash Powdery Mildew}  & \text{d. Potato Blight} 
\\
\\

\includegraphics[width=3.5 cm, height=4cm]{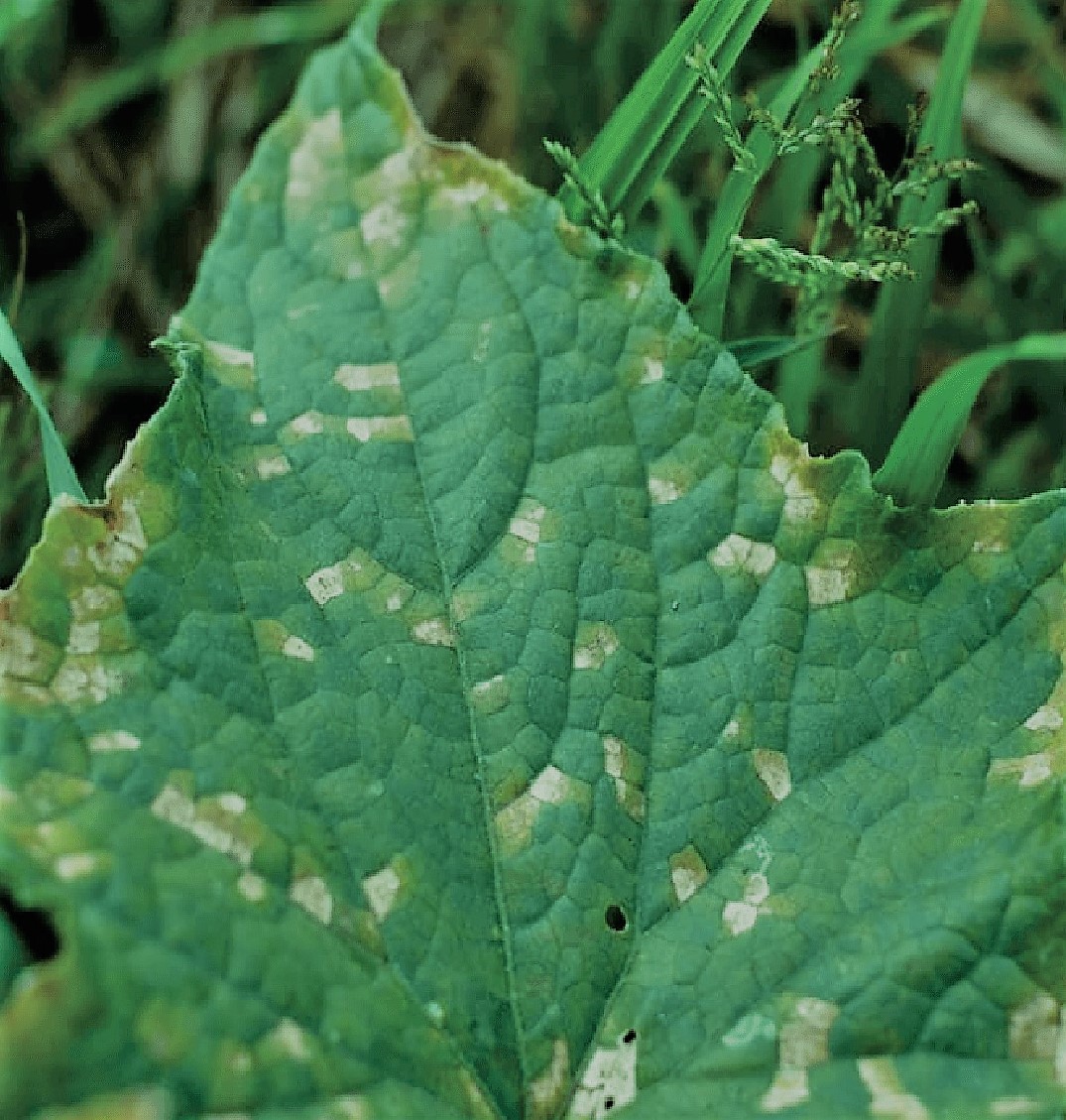} & \includegraphics[width=3.5 cm, height=4cm]{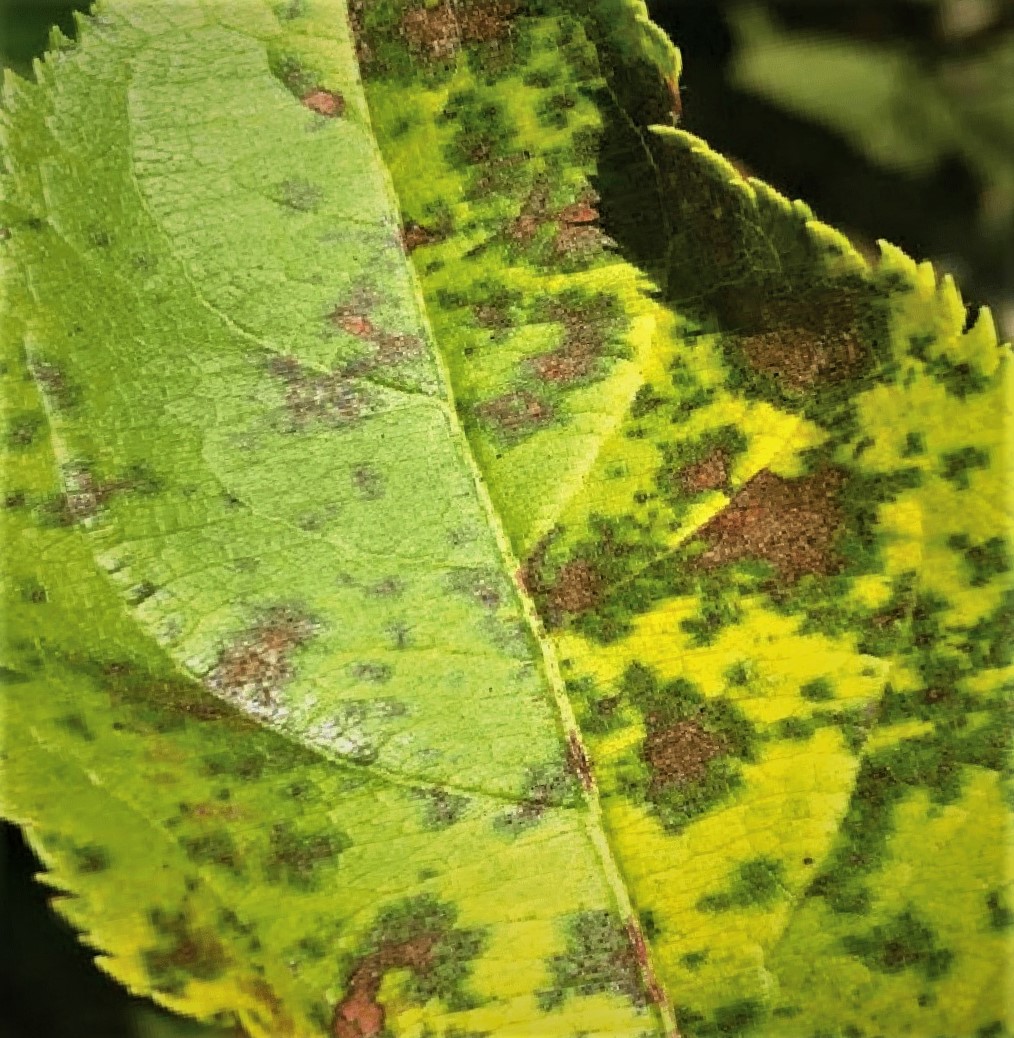}
\\

\text{e. Cucumber Mosaic}  & \text{f. Apple Marssonina Blotch} \\
\\

\includegraphics[width=3.5 cm, height=4cm]{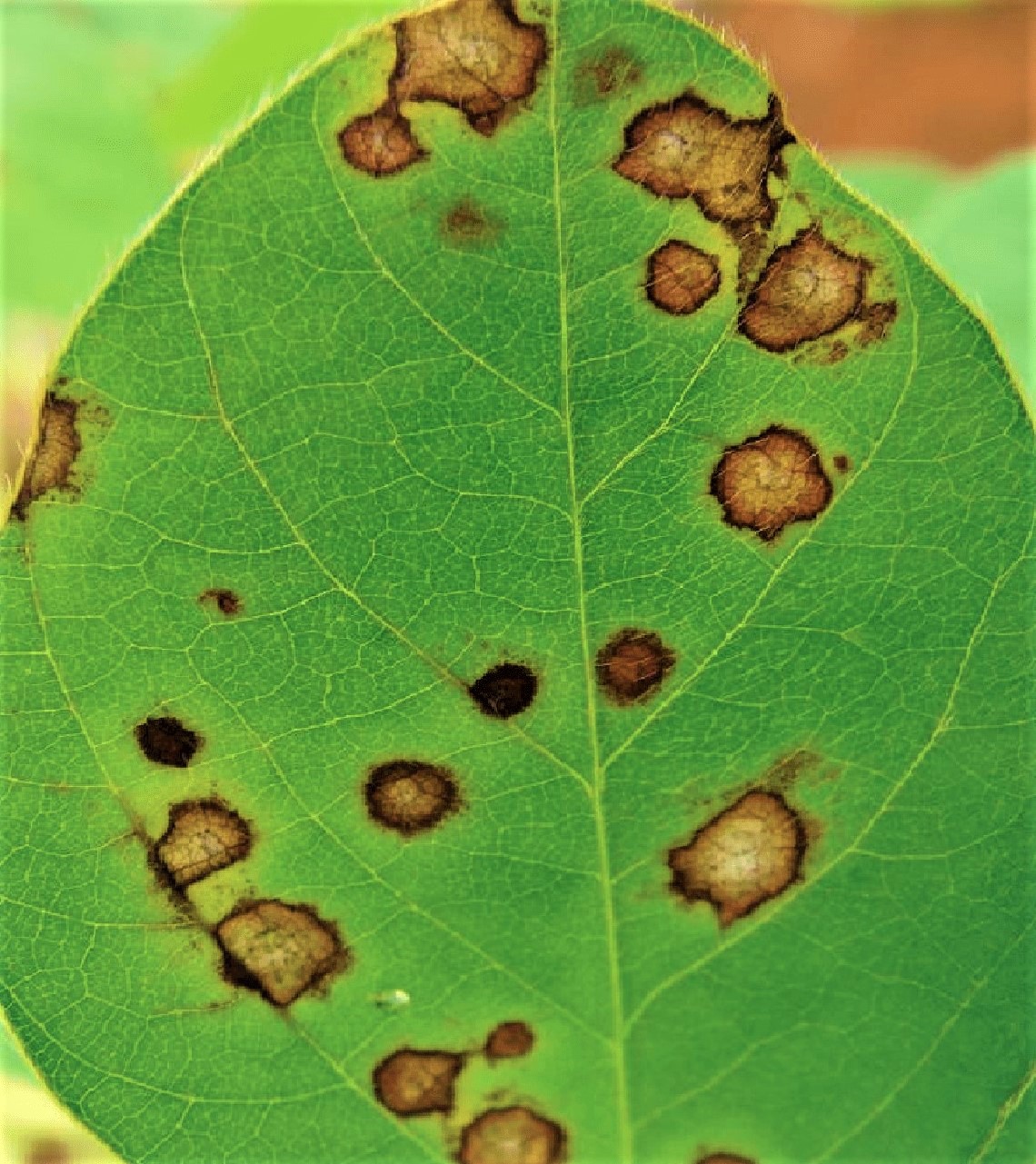} & 
\includegraphics[width=3.5 cm, height=4cm]{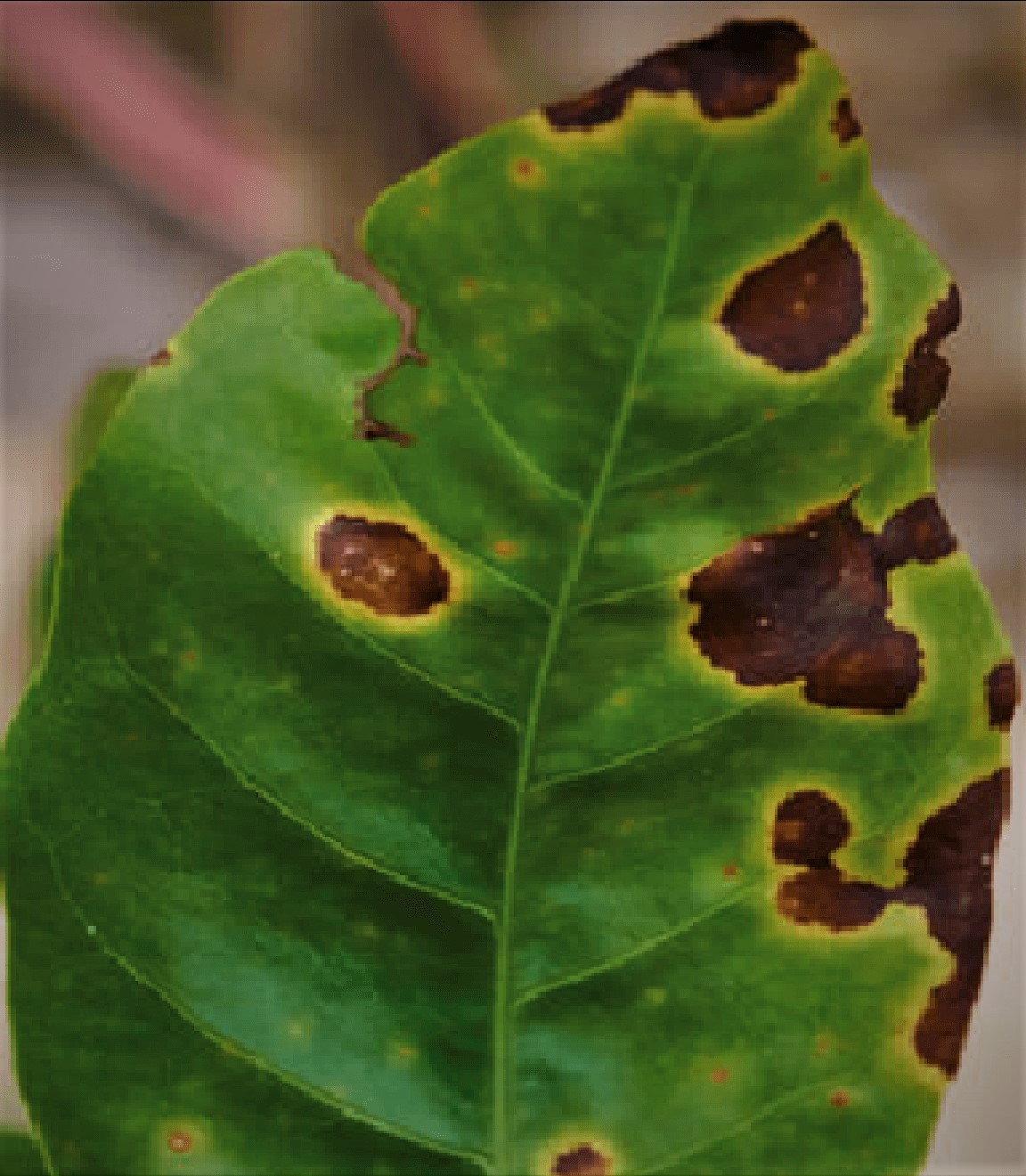}
\\

\text{g. Soybean Frogeye Spot}  & \text{h. Coffee Rust} 
\\
 
\end{tabular} 
\caption{Diseases in various plant leaves.} 
\label{table:1} 
\end{figure}

\begin{table*}[!b]
  \caption{Avilable datasets and their respective leaf diseases}
  \small
  \centering

    \begin{tabular}{ p{5cm} p{3cm} p{2.5cm} l  }
    \hline\\
    \textbf{Dataset Name} & \textbf{Plant Type} & \textbf{\#Images} & \textbf{Leaf Diseases} \\ \\
    \midrule
                                  &        &       & Bacterial Blight\\
    Cassava Leaf Disease \cite{noauthor_cassava_nodate}    & Casava & 21,400 & Brown Streak \\
                                  &        &        & Green Mottle \\
                                  &        &        & Mosaic \\ 
    \midrule
                                  &        &        & Bacterial Spot\\  
                                  &        &        & Early Blight\\
                                  &        &        & Late Blight\\
                                  &        &        & Leaf Mold \\
    Tomato Leaf Disease \cite{noauthor_tomato_nodate}    & Tomato  &11,000  & Septoria Leaf Spot\\
                                  &        &        & Spider Mites\\
                                  &        &        & Target Spot \\
                                  &        &        & Yellow Leaf Curl Virus\\
                                  &        &        & Mosaic Virus\\ 
    \midrule
                                  &        &       & Common Rust\\
    Corn and Maize Leaf Disease \cite{noauthor_corn_nodate}    & Corn    & 4,100 & Gray Leaf Spot \\
                                  & Maize  &      & Blight \\ 
    \midrule
                                  & Apple        &        & \\  
                                  &  Blueberry   &        & Powdery Mildew\\
                                  &  Cherry      &        & Scab\\
                                  &  Corn        &        & Black Rust \\
                                  & Grape        &        & Gray Leaf Spot\\
                                  &   Orange     &        & Common Rust\\ 
    Plant Village  \cite{noauthor_plantvillage_nodate}    & Peach        & 54,303  & Black Measles\\
                                  &  Bell Pepper &        & Huanglongbing\\
                                  &  Potato      &        & Northern Leaf Blight\\
                                  & Raspberry    &        & Bacterial Spot\\
                                  & Soybean      &        & Early Blight\\
                                  & Squash       &        & Late Blight\\
                                  &  Tomato      &        & \\
                                  & Strawberry   &        & \\ 
   \midrule
                              &         &       & Rust\\
    Plant Pathology 2020  \cite{noauthor_plant_nodate-1}    & Apple   & 3,650 & Multiple Diseases\\
                              &         &      & Scab\\
    \midrule
                                 &        &        & Complex\\  
                                  &        &        & Frog-Eye Leaf Spot\\  
                                  &        &        & Powdery Mildew\\
                                  &        &        & Rust\\
                                  &        &        & Rust Complex \\
    Plant Pathology 2021 \cite{noauthor_plant_nodate}    & Apple  & 23,000  & Scab\\
                                  &        &        & Rust-Frog-Eye Leaf Spot\\
                                  &        &        & Powdery Mildew Complex \\
                                  &        &        & Frog-Eye Leaf Spot Complex\\
                                  &        &        & Scab-Frog-Eye Leaf Spot\\
                                  &        &        & Scab-Frog-Eye Leaf Spot Complex\\ 

    \midrule
    \end{tabular}
  \label{tab:1}
\end{table*}

\item{\textbf{Black Spot:}}
Black Spot is another disease that falls into the category of fungal infection, which causes round black spots on the upper side of the leaves (Fig. 1.b). It occurs due to a long period of wet weather or when the plant leaves are kept moist for more than 6 hours. 

\item{\textbf{Powdery Mildew:}}
This fungal disease generally occurs in those plants, usually in shaded areas. It is one of the most easily identifiable diseases, as the infected leaves will have a white powdery substance on the upper surface (Fig. 1.c). This disease generally spreads during humid and low soil moisture conditions. 

\item{\textbf{Blight:}}
Blight is one of the most lethal plant diseases
(Fig. 1.d). In the 1840s potato famine, around one
million people died due to the blight that infected
potato crops across Europe. This disease can only
spread in warm and humid conditions. It falls in the category of fungal disease that spreads through wind-borne spores. 

\item{\textbf{Mosaic:}}
The mosaic virus destroys the plant at its molecular level. It is commonly found  in tomatoes, tobacco, and other horticultural plants. Leaves infected with the mosaic virus will have stripes (Fig. 1.e) of a yellowish and whitish color.

\item{\textbf{Marssonina Blotch:}}
It is a type of fungal disease caused by Marsonina Caronaria. This disease generally occurs in regions that have high rainfall. Infected leaves will have circular dark green patches on the upper surface (Fig. 1.f), which turn to dark brown in the worst scenario.

\item{\textbf{Frogeye Spot:}}
It also fell in the category of fungal disease caused by Cercospora Sojina. In early spring, purple spots on the upper surface of the leaves are a sign of infection. These lesions then grow in a circular ring that remains purple, but the area under the circular ring turns brownish (Fig. 1.g), resembling a frog eye.

\item{\textbf{Rust:}}
Rust is another easily identifiable fungal disease. As the name suggests, it has brownish rusty spots all over the leaves (Fig. 1.h). This disease generally spreads during wet weather in early spring. Mostly found on apples, roses, tomatoes, etc. 
\end{itemize}

\subsection{\centering\normalfont{\textbf{Available Datasets}}}

\subsubsection{\textbf{Cassava Leaf Disease Dataset}}
Cassava is a woody shrub that is native to South America. This dataset \cite{noauthor_cassava_nodate} contains 21,400 images of Cassava plant leaves infected with four different diseases, as mentioned in Table I.

\subsubsection{\textbf{Tomato Leaf Disease Dataset}}
This dataset \cite{noauthor_tomato_nodate} contains images of diseased tomato plant leaves. The dataset is divided into two sets, the first containing 10,000 images of tomato leaves for training and the second containing 1000 images for testing. As mentioned in Table I, these tomato plant leaves are infected with nine diseases.

\subsubsection{\textbf{Corn and Maize Leaf Disease Dataset}}
This dataset \cite{noauthor_corn_nodate} includes 4100 images of diseased Corn and Maize plant leaves infected with three different diseases, as mentioned in Table I.

\subsubsection{\textbf{Plant Village Dataset}}
The Plant Village Dataset \cite{noauthor_plantvillage_nodate} consists of 54,303 images of healthy and unhealthy leaves divided into 38 categories by species and disease. The dataset contains leaves of 14 different plant species with 11 diseases, as mentioned in Table I.

\begin{table*}[t]
  \caption{Conventional plant leaf disease detection and classification techniques.}
  \centering
  \begin{tabular}{l c l l l} \toprule
    \textbf{Paper} & \textbf{Year} & \textbf{Dataset} & \textbf{Technique} & \textbf{Accuracy} \\ 
    
    \midrule
       
    Gavhale et al. \cite{gavhale_unhealthy_2014}    & 2014 & Self-acquired Citrus Leaf Dataset  & SVM-RBF and SVM-POLY & 95{$\%$} and 96{$\%$} \\

    Rothe et al. \cite{rothe_cotton_2015}    & 2015 & Self-acquired Cotton Leaf Dataset & KNN, K-Means, and Texture based Classifier & 80{$\%$} \\ 

    Joshi et al. \cite{joshi_monitoring_2016}    & 2016  & Self-acquired Rice Leaf Dataset & KNN and Minimum Distance Classifier  &  87.02{$\%$} and 89.23{$\%$} \\

    Zhang et al. \cite{zhang_leaf_2017}   & 2017 & Cucumber Leaf Dataset  & Back Propagation based Classifier & 85.7 {$\%$} \\

    Ramesh et al. \cite{ramesh_plant_2018}     & 2018 & Self-acquired Papaya Leaf Dataset & Image Processing and SVM & 70{$\%$} \\
    
    \bottomrule
\end{tabular}
  
  \label{tab:2}
\end{table*}

\subsubsection{\textbf{Plant Pathology 2020 Dataset}}
The Plant Pathology 2020 Dataset \cite{noauthor_plant_nodate-1} contains 3,650 RGB images of 3 different types of apple leaf diseases, as mentioned in Table I.

\subsubsection{\textbf{Plant Pathology 2021 Dataset}}
The Plant Pathology 2021 dataset \cite{noauthor_plant_nodate} contains over 23,000 high-quality RGB images of apple leaves infected with 11 different expert-annotated diseases, as mentioned in Table I.\newline \\All the above-surveyed datasets are well labeled and available on 'Kaggle' in an organized manner.\\

\subsection{\centering\normalfont{\textbf{{LITERATURE REVIEW}}}}

In this section, we embark on a holistic exploration of diverse techniques employed for the classification of multi-class leaf diseases, placing particular emphasis on conventional approaches, deep learning-based methods, and the advancements in computer vision transformers.

\subsubsection{\textbf{Conventional methods}}
Conventional methods for agricultural leaf disease classification typically involve feature extraction and traditional classification algorithms like decision trees, support vector machines (SVM), k-nearest neighbors (KNN), random forests, and naive Bayes classifiers. These algorithms leverage the extracted features to learn patterns and classify the leaf images into different disease categories. Gavhale et al. \cite{gavhale_unhealthy_2014} 2014 proposed a solution that used the conventional Image processing approach to detect unhealthy regions of Citrus leaf that includes five steps  - preprocessing, enhancing leaf image,  segmentation, feature extraction, and disease classification. The dataset was created by manually collecting leaf images of grapefruit, lemons, limes, and oranges infected with diseases such as citrus canker and anthracnose. A total of 200 images were used for training and 50 images for final testing. Feature vectors were extracted and fed to two versions of the Support Vector Machine, SVM-RBF, and SVM-POLY, to classify leaf diseases. Following the conventional approach, Rothe et al. \cite{rothe_cotton_2015} devised a Cotton Leaf Disease Identification and Classification solution using a pattern recognition system. They manually collected cotton leaf images from the Central Institute of Cotton Research fields. The leaf images were infected with three diseases: Bacterial Blight, Myrothecium, and Alternaria. The researchers used an active contour model to segment leaf images, and then image features were extracted to train the neuro-fuzzy inference model. Extracted features are then fed into the Back-Propagative neural network for disease classification.

Joshi et al. \cite{joshi_monitoring_2016} proposed a solution for rice disease control and classification with the help of image processing and machine learning-based classifiers. The researchers used self-acquired rice leaf images of four rice diseases: bacterial-blight, blast,  brown-spot, and sheath-rot. Different features like the shape and the color of the infected region of the leaf were extracted and fed into the two different classifiers: Minimum Distance Classifier (MDC) and K-Nearest Neighbor classifier (KNN) for the disease classification.

Zhang et al. \cite{zhang_leaf_2017} proposed a novel solution for cucumber disease recognition and classification using sparse representation. The researchers proposed a simple three-stage procedure:  segmenting diseased leaf images (using K-means clustering), extracting features from lesion information (shape and color), and disease classification. In addition, practical experiments were conducted and compared for performance analysis using four different feature extraction-based plant leaf disease recognition methods. The cucumber leaf images dataset was manually collected from Northwest A\&F University, which consisted of seven different types of leaf diseases: corynespora cassiicola, bacterial angular, gray mold, scab, downy mildew, anthracnose, and powdery mildew. Finally, the extracted features dictionary was created from diseased leaf images, which the sparse model solver then processed to obtain classification results.

Ramesh et al. \cite{ramesh_plant_2018}
proposed another technique using machine-learning-based classifiers for Papaya leaf disease detection and classification. The proposed solution includes the following steps: dataset creation, feature extraction, classifier training, and disease classification. The dataset is created using manually collected papaya leaf images from the local farms, to which a feature 
extracting technique called Histogram of an Oriented Gradient (HOG) was applied. The extracted features are then fed into various classifiers such as Logistic Regression, SVM, KNN, CART, Random Forests, and Naive Bayes. Based on performance evaluation, the researchers concluded that the Random Forest classifier outperformed others in accuracy.

\subsubsection{\textbf{Deep Learning methods}}
Based on the above study, we can note that conventional methods for leaf disease classification heavily rely on manual feature engineering, which requires domain expertise and may limit their ability to capture complex patterns and variations in the data. However, with the advancements in deep learning, more modern approaches based on deep neural networks have gained popularity due to their ability to automatically learn hierarchical representations from raw data, potentially improving classification accuracy in leaf disease detection tasks. In 2016, Mohanty et al. \cite{mohanty_using_2016}
proposed a solution that uses deep convolutional neural networks to detect disease in plant leaves by focusing on two popular architectures, AlexNet and GoogLeNet. They used the Plant Village Dataset consisting of 26 different leaf diseases, as shown in Table I. Before feeding the dataset to the black-box model, the images were first downscaled to 256 × 256 pixels. Also, the Plant Village dataset was divided into three different types. The first type was trained with RGB-colored images, followed by grayscaled and segmented image datasets, where all the unimportant background information was masked, which may introduce some bias in the dataset due to the data collection process. Baranwal et al. \cite{baranwal_deep_2019}
On the other hand, we proposed a deep-learning-based approach for detecting apple leaf disease using the GoogLeNet architecture. For better and more accurate
detection of the infected region in the leaf, the method also
employed image augmentation and transformation techniques
such as downsizing the image size to 60 × 60 pixels, rotation,
width shifts, height shifts, and horizontal and vertical flips. Singh et al. \cite{singh_multilayer_2019}
used a Multilayer-Convolutional Neural Network (MCNN) model that uses AlexNet architecture for the disease classification of the Mango leaves infected with Anthracnose fungal disease. They used two different datasets. The first is the self-acquired Mango leaves dataset, and the other is the Plant Village dataset mentioned in Table I.

\begin{table*}[t]
  \caption{Deep learning-based plant leaf disease detection and classification.}
  \centering
  \begin{tabular}{l c l l l} \toprule
    \textbf{Paper} & \textbf{Year} & \textbf{Dataset} & \textbf{Technique} & \textbf{Accuracy} \\ 
    
    \midrule
     
    Mohanty et al. \cite{mohanty_using_2016}    & 2016  & Plant Village Dataset & AlexNet and GoogLeNet  & 85.53{$\%$} and 99.34{$\%$} \\ 

    De Luna et al. \cite{de_luna_automated_2018}    & 2018  & Self-acquired Tomato Leaf Dataset & Faster R-CNN & 91.67 {$\%$} \\

    Singh et al. \cite{singh_multilayer_2019}    & 2019  & Mango Leaf and Plant Village Dataset & Multilayer Convolutional Neural Network (MCNN)  & 97.13{$\%$} \\

    Baranwal et al. \cite{baranwal_deep_2019}    & 2019   & Plant Village Dataset & CNN & 98.42{$\%$} \\

    Tiwari et al. \cite{tiwari_potato_2020}    & 2020  & Plant Village Dataset  & Logistic Regression Classifier & 97.8 {$\%$} \\

    Srinidhi et al. \cite{srinidhi_plant_2021}    & 2021  & Apple Leaf Dataset from CIDA & EfficientNet and DenseNet   & 99.8{$\%$} and 99.75{$\%$} \\

    Sujatha et. al. \cite{sujatha_performance_2021}    & 2021  & Self-acquired Citrus Leaf Dataset & Various ML and DL (VGG16)  & 89.5{$\%$} \\

    Singh et al. \cite{singh_hybrid_2022}    & 2022 & Plant Village Dataset & Bayesian Optimized SVM, and Random Forest & 96.1{$\%$} \\
    
    \bottomrule
\end{tabular}
  
  \label{tab:2}
\end{table*}

With the availability of better deep-learning classifiers models,
De Luna et al. \cite{de_luna_automated_2018}
used a Faster-Region-based CNN to identify tomato leaf diseases. A motor-controlled image-capturing robot was made to capture images from four different angles of the tomato plant to detect and recognize leaf diseases. The study focused on AlexNet architectures and disease recognition networks. Similarly, Srinidhi et al. \cite{srinidhi_plant_2021}
used two more efficient and advanced Deep-CNN models, EfficientNet and DenseNet, to detect the infected leaves and accurately classify them into various diseases mentioned in Table I. Also, the authors have compared these models with other
CNN models, like AlexNet, VGG, and GooLeNet, perform low compared to the proposed models due to information loss (spatial and rotational). They used Apple Leaf Dataset from Cornell Initiative for Digital Agriculture (CIDA). The authors improved the dataset by employing data augmentation and image annotation techniques such as Flipping, Blurring, and Canny Edge Detection. On the other hand, Sujatha et al. \cite{sujatha_performance_2021} took a different approach, combining both Machine Learning and Deep Learning to solve the Citrus plant leaf disease detection problem. The dataset was collected manually under the guidance of experts and Citrus Research Centre, Government of Punjab. In addition, machine learning and deep learning approaches were compared (Models used: SVM, Random Forest, Stochastic Gradient Descent, Inception-v3, VGG-16, and VGG-19). They found that deep learning methods outperformed machine learning approaches by a wide edge.

Tiwari et al. \cite{tiwari_potato_2020} proposed a comparison-based disease diagnosis and classification solution using transfer learning. Plant Village Dataset extracts relevant features from the potato leaf dataset, which is then fed into a pre-trained VGG19 model for fine-tuning. The authors then used multiple classifiers to obtain the two most optimal results. Singh et al. \cite{singh_hybrid_2022} 
 methodologies and compared the simulated results for performance evaluation. The first method used data augmentation techniques on the leaf image dataset to extract deep features using CNN. The extracted features were then classified into disease classes using a Bayesian optimized support vector machine classifier. Data preprocessing techniques were first applied to the image dataset in the second method. Texture and color features were extracted using the histogram of oriented gradient (HoG) and color moments. These extracted features were then grouped to form hybrid features, which were then selected using swarm optimization before classification with a random forest classifier to obtain simulation results. The researchers have also explored the idea of detecting plant leaf diseases in the compressed domain using deep learning models \cite{javed_CDomain}\cite{javed_t2CIGAN}.

\subsubsection{\textbf{Vision Transformer methods}}
Transformers, a specific class of deep learning architecture, have garnered substantial recognition, particularly in natural language processing (NLP) and computer vision. Vaswani et al. introduced transformers in 2017 as an alternative to recurrent neural networks (RNNs) for sequence modeling tasks, showcasing their exceptional ability to capture long-range dependencies. Their effectiveness has been widely demonstrated in machine translation, language generation, and text classification tasks. A distinguishing feature of transformers is their utilization of a self-attention mechanism, enabling the model to weigh the significance of individual elements within a sequence during processing. This attention mechanism facilitates efficient modeling of relationships and dependencies among the elements. Comprising encoder and decoder components that can be stacked, the transformer architecture allows for the creation of deeper models with enhanced representational capacity.

Although initially popularized in NLP, transformers have succeeded remarkably in diverse fields, including computer vision. Vision transformers (ViTs) adapt the transformer architecture for image recognition tasks. Similar to text-based transformers, ViTs leverage self-attention mechanisms to model relationships between different patches within an image. By dividing an image into smaller patches and treating them as input tokens, ViTs process visual data in a sequential manner, akin to sequences of text. A significant breakthrough in vision transformers occurred in 2020 with the introduction of the "Vision Transformer" model by Dosovitskiy et al. \cite{dosovitskiy_image_2021}, showcasing its competitive performance in image classification tasks and affirming the transformers' capacity to handle visual data effectively. Hence, we present the study showcasing remarkable advancements made by transformers in addressing the complex challenge of multi-class plant leaf disease classification.

Zhenghua Zhang et al. \cite{zhang_swin-transformer_2021} proposed a rice disease 
identification method based on the Swin Transformer, incorporating sliding window 
operation and a hierarchical design. The model architecture shares similarities with CNN, 
featuring stages and repeating units within each stage. The paper addresses the challenge of significant visual entity variations and the potential limitations of visual transformers in diverse scenes. To develop their approach, the authors gathered a dataset comprising 6298 images of infected rice leaves from various sources, including Google, Baidu, and Kaggle. The dataset consists of five classes: rice stripe (1026 images), rice blast (1097 images), rice false smut (1770 images), brown spot (1525 images), and sheath blight (880 images). The authors assert that their model outperforms traditional machine learning models by 4.1\%, indicating its efficacy in enhancing rice leaf disease recognition compared to conventional approaches.

\begin{table*}[t]
  \caption{Vision Transformer based plant leaf disease detection and classification.}
  \centering
  \begin{tabular}{l c l l l} \toprule
    \textbf{Paper} & \textbf{Year} & \textbf{Dataset} & \textbf{Technique} & \textbf{Accuracy} \\ 
    
    \midrule
    
    Zhenghua Zhang et al. \cite{zhang_swin-transformer_2021}    & 2021  & Custom Dataset & Swin Transformer  & 95.2{$\%$}\\ 

    Fengyi Wang et al. \cite{wang_practical_2022}    & 2022  & Self-acquired Cucumber Leaf Dataset & Improved Swin Transformer & 98.97 {$\%$} \\

    Hamoud Alshammari et al. \cite{alshammari_olive_2022}    & 2022  & Self-Acquired Olive Leaves Dataset & CNN + Vision Transformer  & 97.0{$\%$} \\

    Amer Tabbakh et al. \cite{tabbakh_deep_2023}    & 2023   & PlantVillage and Wheat Dataset & Vision Transformer & 99.86{$\%$} \\

    Huy-Tan Thai et al. \cite{thai_formerleaf_2023}    & 2023  & Cassava Leaf Disease Dataset & FormerLeaf with LeIAP and SPMM  & 96.8{$\%$}\\ 

    Bin Yang et al. \cite{yang_identifying_2023}    & 2023  & AI Challenger 2018 Dataset & Triple-Swin Transformer & 99.0{$\%$} \\

    Sheng Yu et al. \cite{yu_inception_2023}    & 2023  & Plant-Village, Ibean, AI2018, and PlantDoc Dataset  & Inception + Vision Transformer & 99.94{$\%$} \\

    Li Ma et al. \cite{ma_maize_2023}    & 2023 & Plant-Village Dataset & YOLOv5 + Swin Transformer & 95.2{$\%$} \\

    \bottomrule
\end{tabular}
  
  \label{tab:3}
\end{table*}

Fengyi Wang et al. \cite{wang_practical_2022} proposed an innovative approach for identifying cucumber leaf diseases using an improved variant of the Swin Transformer (SwinT). They addressed the challenge of information loss among patches generated by the patch partition layer in the original SwinT by introducing step-wise small patch embeddings. This modification enhanced the feature extraction capability without increasing the model's parameters. The authors developed a leaf extraction module by integrating the improved SwinT and Grad-CAM into a Generation Adversarial Network (GAN) to augment the disease dataset and focus on the leaf region of images with complex backgrounds. This integration led to the creation of STA-GAN (SwinT-based and Attention-guided GAN), which selectively generated disease spots only within the leaf region. The dataset used in their study comprised 4740 cucumber leaf images captured by the authors under natural light conditions from farms in Dawei Town, Hefei City, Anhui Province, China. It included 1800 healthy and 2940 diseased leaf images representing Anthrax, Downy mildew, Powdery mildew, and Target spot diseases. The generated dataset was employed along with transfer learning using the proposed network to train the recognition model for cucumber leaf diseases. The authors also compared their findings with LeafGAN \cite{cap_leafgan_2022}, an image-to-image translation system utilizing a custom attention mechanism. LeafGAN performed transformations on healthy leaf images to generate a diverse range of diseased images, which were used as an augmentation technique to enhance plant disease diagnosis performance. According to the testing results, STA-GAN produced higher-quality images than LeafGAN, 
demonstrating a superior performance improvement of 2.17\%.

A hybrid deep learning model for olive leaf disease classification was proposed by Hamoud Alshammari et al. \cite{alshammari_olive_2022}. Their model combines the convolutional neural network (CNN) and vision 
transformer architectures to leverage their respective strengths. The entire architecture can be divided into three steps. Firstly, a preprocessing step is applied to the dataset, utilizing a noise filter algorithm to remove noise and enhance image quality. A data enhancement procedure is conducted to augment the dataset further. The preprocessed dataset is then fed into the hybrid model to extract essential features from the infected regions of the leaf images. The third and final step involves image classification, incorporating a pooling layer and dropout mechanism to address the issue of overfitting. Before the final stage, a SoftMax layer is applied for classification. The authors employed a dataset of 3400 self-acquired olive leaf images for their research. This dataset was categorized into three distinct classes: healthy leaves, leaves infected with oculus olearius, and leaves infected with olive peacock spot. To evaluate the performance of their hybrid model, the authors compared it with other deep learning models, namely VGG16, VGG19, and ViT. The results demonstrated that the hybrid learning model achieved significantly higher accuracy rates compared to the other models.

Amer Tabbakh et al. \cite{tabbakh_deep_2023} introduced a novel hybrid model, 
TLMViT, for plant disease classification. TLMViT combines Transfer Learning with a Vision 
Transformer approach, utilizing a four-stage architecture. The model includes data 
acquisition, image augmentation, and leaf feature extraction through two consecutive phases: 
Initial features were extracted using a pre-trained model, and in-depth features were extracted using Visual Transformer. In the final stage, an MLP classifier is employed for the classification task. The authors conducted experiments using five pre-trained-based models followed by ViT individually, comparing them with TLMViT. The results revealed improved performance with the transfer learning-based TLMViT model compared to the pre-trained-based architectures. The Plant-Village dataset was analyzed, and TLMViT exhibited exceptional performance for classifying plant leaf disease.

Huy-Tan Thai et al. \cite{thai_formerleaf_2023} introduced a novel leaf disease 
detection model called FormerLeaf, based on the transformer architecture. Additionally, the authors proposed two performance optimization techniques, namely LeIAP (Least Important Attention Pruning) and SPMM (Sparse Matrix-Matrix Multiplication) algorithms, to improve the model's efficiency. The LeIAP algorithm selectively prunes unimportant attention heads in each layer. In contrast, the SPMM algorithm utilizes sparse matrix-matrix multiplication to calculate matrix correlation, effectively reducing the model's complexity and training time by 10\% without compromising performance quality.
The authors evaluated their model using the Cassava Leaf Disease Dataset, which comprises 
21,367 labeled images. The results indicate that their model achieves a remarkable reduction in model size, shrinking it to only 28\%. Moreover, it significantly improves  training and inference speed, with a 10\% and 15\% increase, respectively. These findings highlight the effectiveness of FormerLeaf in efficiently detecting leaf diseases, making it a promising solution in the field.

In a quest for advanced disease classification and severity detection, Bin Yang et al. \cite{yang_identifying_2023} introduced a Triple-Swin Transformer Classification (TSTC) model. This transformative approach leverages the power of swin transformers, pushing the boundaries of disease analysis. The TSTC architecture comprises three remarkable modules, each serving a specific purpose. The multitasking feature extraction module harnesses the strength of a triple-branch network to extract preliminary features dedicated to disease and severity classification. These preliminary features are then seamlessly combined through the feature fusion module, utilizing the cutting-edge CBP technique to enhance their quality. To extract even more discriminative features, the deep supervision module supports the hidden layers of the TSTC network. To test their model's prowess, the authors turned to the AI Challenger 2018 dataset, a vast collection of 31,718 leaf disease images. Their TSTC model boasts an overall accuracy of 99.0\% for disease classification, ensuring accurate identification, and an impressive 88.73\% for severity classification, enabling precise assessment of the disease's impact.

In their research, Sheng Yu et al. \cite{yu_inception_2023} propose a ground-breaking end-to-end 
deep learning model called Inception Convolutional Vision Transformer (ICVT). This novel 
model combines the effectiveness and efficiency of the inception convolution framework 
with the Transformer architecture. It excels in terms of both network parameters and 
classification accuracy. The ICVT model not only captures local spatial features of 
surrounding tokens but also emphasizes the learning of high-level information crucial for 
plant disease identification. Incorporating the inception architecture and cross-channel feature learning further enhances the richness of information, particularly benefiting fine-grained feature learning within the model. The authors evaluate the ICVT model using four distinct datasets: Plant-Village, Ibean, AI2018, and PlantDoc. Through their analysis, the proposed model surpasses previous convolutional and vision transformer-based models, achieving unprecedented accuracy rates. Specifically, it achieves an accuracy of 99.94\% on the Plant-Village dataset, 99.22\% on Ibean, 86.89\% on AI2018, and 77.54\% on PlantDoc. The experimental results demonstrate the superiority of the ICVT model over existing approaches, solidifying its preponderance in plant disease identification.

Li Ma et al. \cite{ma_maize_2023} have presented an innovative model known as CTR-YOLOv5n for maize leaf disease recognition. This model seamlessly integrates the robust YOLOv5n backbone with two key components, namely the Coordinate Attention (CA) mechanism and the Swin Transformer (STR). By incorporating the CA mechanism into the backbone network, the model's proficiency in identifying maize leaf spots is significantly enhanced, as it allocates greater importance to the spot-related features, thus enabling the effective classification of small leaf spots. In parallel, integrating the Swin Transformer into the larger leaf spot detection head amplifies the model's feature-capturing capabilities. This facilitates the accurate detection of more prominent leaf spots within the image. The analysis was conducted using 4353 maize leaf images from the Plant-Village dataset, consisting of four classes: 1162 healthy leaf images, 1000 maize-blotch diseased leaf images, 1191 maize-rust diseased leaf images, and 1000 maize-gray spot diseased leaf images. The experimental results demonstrate significant enhancements, with the proposed model achieving an average recognition accuracy of 95.2\%. Notably, compared to the memory size of YOLOv5l, the proposed model achieves a remarkable 94.5\% reduction in memory requirements, resulting in a more efficient and compact model.

After conducting an extensive literature review, we found how vision-based transformer models offer advantages over conventional and CNN-based deep learning models in the context of multi-class leaf disease classification.

\begin{figure*}[!b]
\centering
 \makebox[\textwidth]{\includegraphics[width = 19 cm, height = 5 cm]{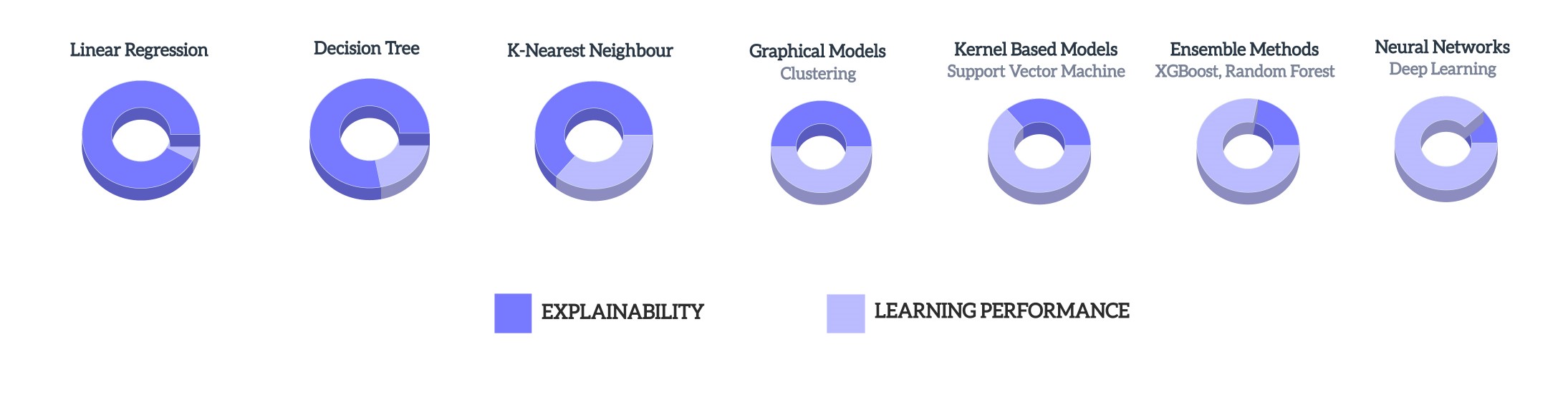}}
\caption{Learning Performance vs. Explainability}
\label{fig:}
\end{figure*}

\section{\textbf{OVERVIEW OF EXPLAINABLE AI (XAI)}}

Nowadays, artificial intelligence is receiving a lot of attention. Almost every research field plans to use AI or converts outdated rule-based technologies to AI-enabled ones. Many of today's rule-based systems, which incorporate deep learning and machine learning technologies, cannot explain how the system operates to its users and the variables involved in crucial decision-making. This gap causes users to lose trust in the technology, eventually stopping them from using the final product application. Furthermore, some AI researchers contend that emphasizing explanations is sometimes unnecessary for AI-based research and is too difficult to accomplish \cite{gunning_xaiexplainable_2019}
. Others contend that explanations should accompany the data provided by AI-based systems to improve human intelligence. Completing this gap may help the industry trust AI systems and open new opportunities for AI-based products and services. 

On the other hand, users need explanations to understand, trustfully, and effectively manage these new artificially intelligent partners, such as law, medicine, agriculture, finance, and defense applications. Providing explanations should be considered an additional layer of human-computer interaction that helps humans to get more value from the services provided by AI-based systems. Recently developed machine-learning techniques that build models from their internal representations are primarily responsible for the field of AI's advancement. Some examples are SVMs, Random Forests (RF), Probabilistic models, and Deep Learning Neural Networks (Black-Box models). These techniques are designed to work without human intervention and can be utilized in multiple situations without customization. An inherent conflict between machine learning-based model performance (e.g., predictive accuracy) and explainability may exist. It is observed that the most accurate methods (e.g., Deep Learning) are often the least explainable, and the least accurate methods (e.g., Decision Trees) are also the most explainable. Fig. 2 provides an example of a hypothetical graph representing some ML technique's performance-explainability trade-off. It can be seen that the higher the accuracy of a method, the lower its explainability. So, to make AI solutions transparent and trustworthy so that they can be presented to users more understandably, a research domain called Explainable Artificial Intelligence (XAI) was introduced.

\begin{figure*}[h]
\centering
\includegraphics[width = 16cm, height = 6.5cm, center]{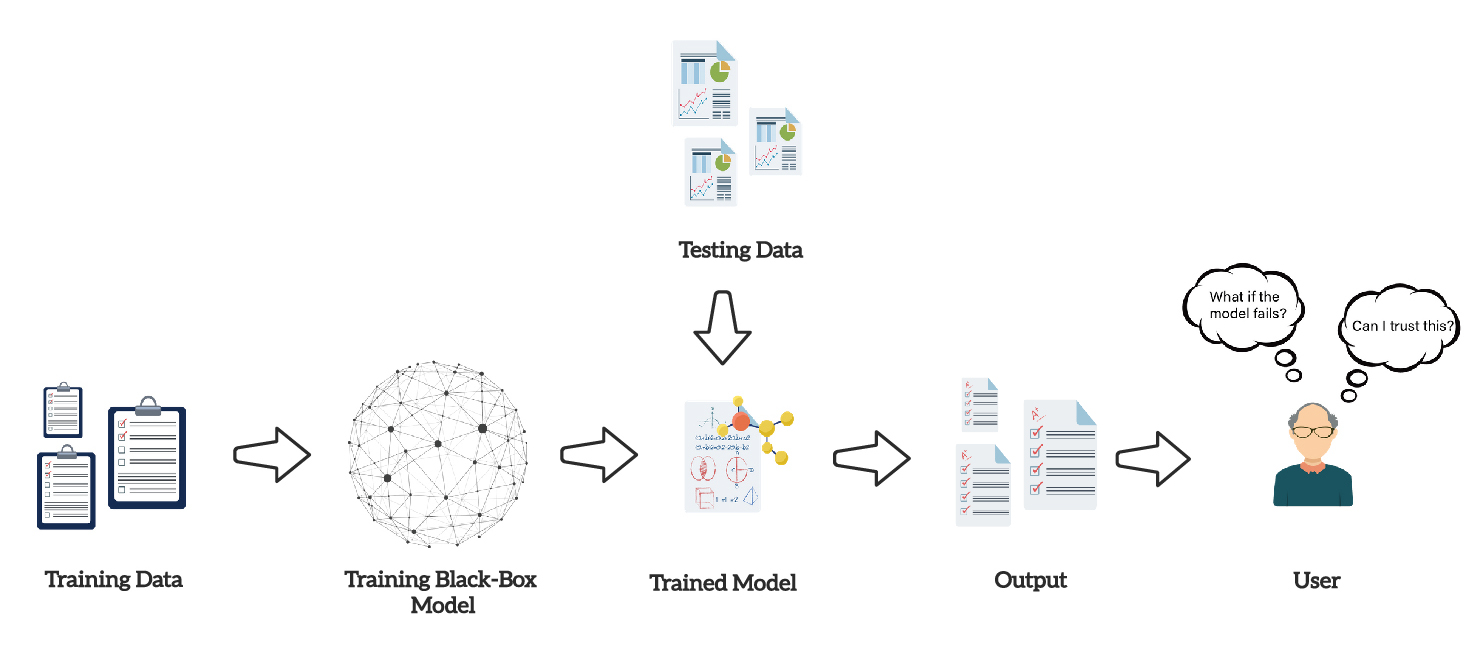}
\caption{Black-Box incorporated into the workflow.}
\label{fig:}
\end{figure*}

\begin{figure*}[!b]
\centering
\includegraphics[width = 17.4cm, height = 8.6cm, center]{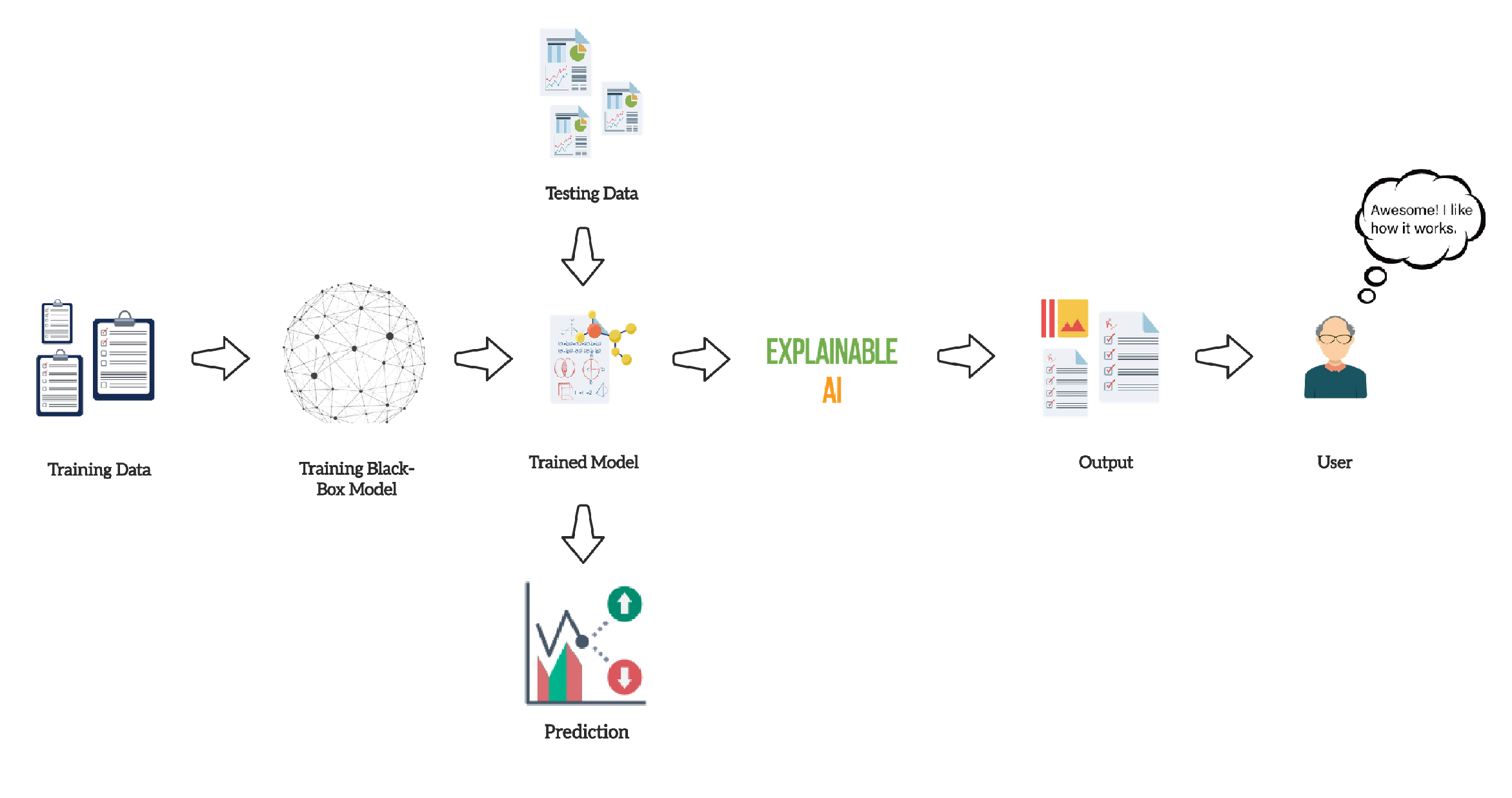}\\
\caption{Explainable AI incorporated into the workflow.}
\label{fig:}
\end{figure*}

\subsection{\centering\normalfont{\textbf{What is XAI?}}}
Artificial intelligence is used in various fields, from autonomous vehicles to medical diagnosis (AI). Users without technical backgrounds seek ways to comprehend the systems they employ so they can have faith in the system's decisions. Looking more closely, critical industries like defense, healthcare, and safety rely heavily on AI talent. We must therefore consider the likelihood that AI may eventually replace human supervisors in these industries. Demonstrating how the AI successfully reached a particular result or solution is essential.  Also, it should explain how the AI works and give users the means to verify that it does what it claims to do. Explainable Artificial Intelligence (XAI), a branch of Machine Learning, was developed to address this issue. Its goal is to make AI systems transparent and trustworthy so users can trust them. XAI employs ethics in AI to reveal the system's inner workings. This increases trust in the model's judgments by decreasing unconscious biases. The primary goal of an Explainable AI system is to make its actions more understandable to humans by providing valid explanations, which can be accomplished by applying some general principles to create more
efficient and humane AI systems. Fig. 3 illustrates the current workflow of the Black-Box-based learning approach.

Now let us understand the above workflow from a hypothetical example. Assume a severe medical condition where a patient has breathing-related issues, and the doctor puts him on a ventilator. Now the doctor is monitoring the patient's heart rate on the AI-incorporated system, which shows a fall and rise in the heart rate on the monitor screen. The algorithm of that AI-incorporated system (highly accurate) is designed in such a way that for the $i^{\text{th}}$ instant of time, it will show the predicted heart rate of the patient for the next 15 sec based on previous and current heart rate. Again, look at Fig. 3 and relate this problem to the given workflow. In this case, the doctor is the existing user relying on the AI-incorporated system's output. Don't you think this workflow is too risky, as the doctor is depending on the Black-Box AI system, which only shows us the predicted heart rate without giving us the proper explanation of what the internal factors are that are causing the expected heart rate variations?
 
So what we have observed from the above example is that even the results of a highly accurate black-box model are neither trustworthy nor understandable to the user to make conclusive decisions. This puts industries relying on Artificial Intelligence at significant risk. As a result, proper explanations are required for these AI-based predictions, which can be fully trusted, and the users can rely on them. Thus, more and more explainable AI systems need to be incorporated, which should be able to explain their understandings like, "What has it done?", "What is it currently doing?" and "What will it do next?" reveal the salient information responsible for the predicted outcome. The workflow for this Explainable AI-based approach is illustrated in Fig. 4.

\subsection{\centering\normalfont{\textbf{Definition and Taxonomy of XAI}}}

Researchers use the terms explainability and interpretability to explain the concept of Explainable AI. \textbf{Explainability} is an active property of a model that refers to the measures taken by a model to explain the internal functioning of the AI system in a human-understandable form. At the same time, \textbf{Interpretability} (also expressed as transparency) is the passive property of a model that enables one to comprehend how a model arrives at a predicted decision in human terms. Interpretability can be further classified into "Intrinsic" and "Post-hoc" interpretability, as shown in Fig. 5.

\begin{figure}[htbp]
\centering
\includegraphics[width = 9cm, height = 5.2cm, center]{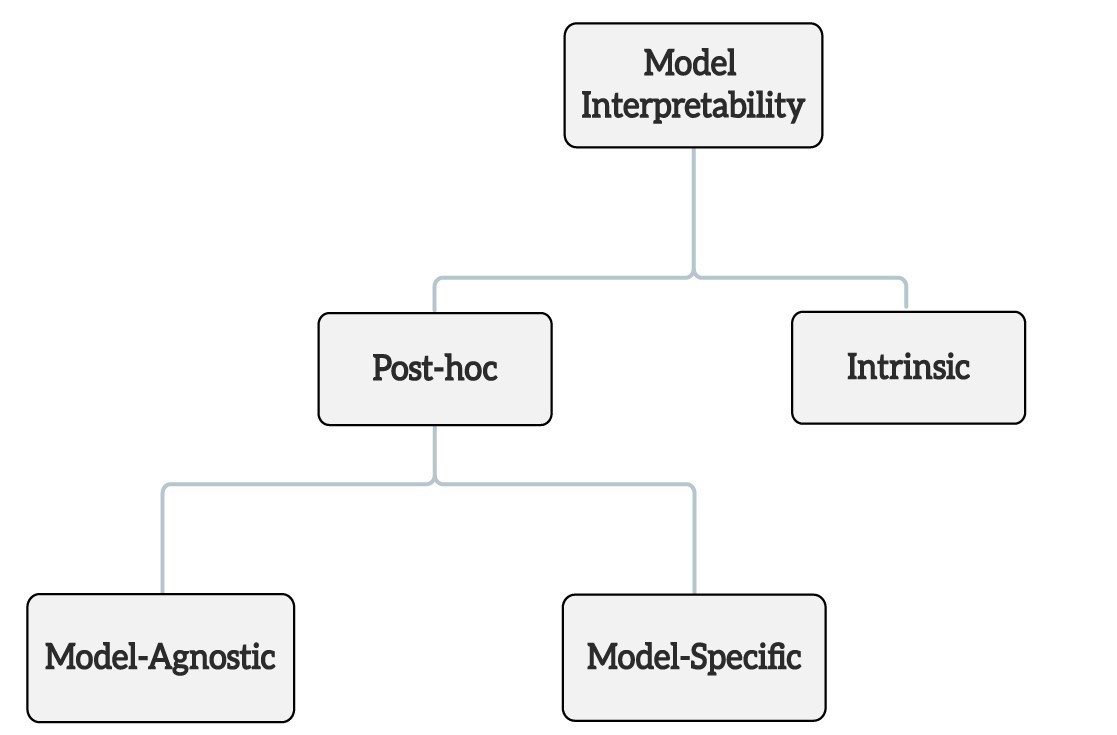}
\caption{Taxonomy of Interpretability.}
\label{fig:}
\end{figure}

\begin{itemize}
    \item\textbf{Intrinsic Interpretability:} Machine learning models with intrinsic interpretability, such as short decision trees and sparse linear models, are considered interpretable due to their simple and less complex structure. In general, we can achieve intrinsic interpretability by limiting the complexity of the machine learning model.

    \item\textbf{Post-hoc Interpretability:} Models like neural networks and gradient boosting are not self-interpretable due to their complex structure. As a result, an external method or technique must be used to explain the facts and feature relationships that drive the model's decision. "Post-hoc interpretability" refers to human-comprehensible explanations obtained after a model's decision-making by an external method. In general, methods that analyze the model after training and decision-making can be used to achieve post-hoc interpretability.

    Post-hoc interpretability is further classified as "model-specific" or "model-agnostic." Model-specific interpretation refers to methods for generating post-hoc interpretation limited to specific model classes. On the other hand, methods that can be used on any machine learning model for acquiring post-hoc interpretations are termed "model-agnostic" interpretations.

\end{itemize}

\subsubsection{\textbf{Scope of Interpretability}}
Another method for comparing model interpretability is to consider the scope. We can determine whether a model is globally or locally interpretable.\\
\begin{itemize}
    \item{Global Interpretability:} A model is globally interpretable if we can fully comprehend its interpretation simultaneously.
    
    \item{Local Interpretability:} A model is locally interpretable if we can understand how a specific decision was reached for the given inputs.
\end{itemize}

\subsubsection{\textbf{White Box and Black Box Models}} AI models can be either white-box or black-box.

\begin{itemize}
    \item{White-Box Models:} White-box models are the more straightforward and least complex models, like linear regression and decision trees, which offer comparably low prediction accuracy compared to black-box models. They are self-explanatory and hence do not necessitate the use of additional capabilities to be explainable.
    
    \item{Black-Box Models:} Black-box models, on the other hand, are more complex and very hard to understand. They are not self-explanatory. Therefore, we must employ various strategies to extract explanations from the model's underlying logic or outputs to make them explainable. Models like neural networks and gradient boosting are examples of Black-box models that provide higher prediction accuracy than White-box models.
\end{itemize}

\subsection{\centering\normalfont{\textbf{Methods of XAI}}}
We've seen how explainable AI can build trust between AI technology and users by offering human-understandable explanations. So let's discuss a few of these XAI approaches.

\subsubsection{\textbf{Explainable AI with SHAP}}
To provide model transparency, Lundberg and Lee devised an explainer SHAP (Shapley Additive Explanations), which uses a game theory-based model agnostic methodology to explain any model. The SHAP delivers three noteworthy features. The first is global interpretability, followed by local interpretability, and finally by SHAP values. SHAP uses game theory's classical Shapley values to connect optimal credit allocation with local explanations.

\subsubsection{\textbf{Explainable AI with LIME}}
To make AI understandable, two sorts of confidence are required: confidence in our model and our forecast. LIME, an explainer, was introduced to earn such trust. It was the first technique to create interest in the explainability domain. LIME is an acronym for "Local Interpretable Model Agnostic Explanations." It is an algorithm capable of faithfully explaining the predictions of any classifier or regression model by approximating them locally using an interpretable model. It alters a single data sample by changing the feature values and then observes the resultant effect on the output. LIME produces a unique set of explanations describing each characteristic's contribution to a prediction for a single sample, which refers to local interpretability.

\subsubsection{\textbf{Explainable AI with Grad-CAM}}
A Microsoft Inc. team developed the InterpretML module, which combines prediction accuracy with model interpretability via an integrated API. If you use scikit-learn as your primary modeling tool, the InterpretML API provides a unified framework API similar to scikit-learn. It extensively uses libraries like Plotly, LIME, SHAP, and SALib and is thus already compatible with other modules. Its Explainable Boosting Machine (EBM) is a simple method based on Generalized Additive Models (GAMs). A GAM's terms are additive, just like those in a linear model, but they do not have to be necessarily linear with the target variable.

\subsubsection{\textbf{Explainable AI with Microsoft’s InterpretML}}
The InterpretML module, created by a Microsoft Inc. team, provides prediction accuracy and 
model interpretability through an integrated API. If you use scikit-learn as your primary modeling 
tool, the InterpretML API provides a unified framework API similar to scikit-learn. It uses many libraries, including Plotly, LIME, SHAP, and SALib, and is thus already compatible with other modules. Its Explainable Boosting Machine (EBM) is an easy-to-understand algorithm 
based on Generalized Additive Models (GAMs). A GAM's terms are additive, just like those in a 
linear model, but they do not have to be linear with the target variable.

\section{\textbf{ XAI IN THE CONTEXT OF PLANT LEAF DISEASE DETECTION}}
This study showed that deep learning and computer vision were essential in plant disease identification and classification. From old traditional methods to machine-learning-based conventional methods and then from conventional methods to deep-learning-based advanced methods, we have upgraded both the methods and approaches at each stage of solving this problem. At present, we can say that deep-learning-based advanced methods are the most efficient way to detect disease in plants. But, we also can't deny that with better performance of deep-learning-based methods, understanding the process behind the solution becomes more difficult. If we compare machine-learning-based conventional methods to deep-learning-based advanced methods, we will find that traditional methods are easier to understand than deep-learning-based methods. Modern methods are more complex due to several layers of nested structures, which makes them very difficult to understand. This is the prime reason that deep-learning models are still considered black-box models. Thus, the decision-making of deep-learning-based methods requires a proper explanation so that, in a real-life scenario, the user (agricultural practitioner) can easily understand the process and the factors behind the classification results and generate trust in the technology. So, in this section, we will examine how researchers have used Explainable AI techniques to address this problem.

\begin{figure*}[h]
\centering
\includegraphics[width = 17.4cm, height = 8.6cm, center]{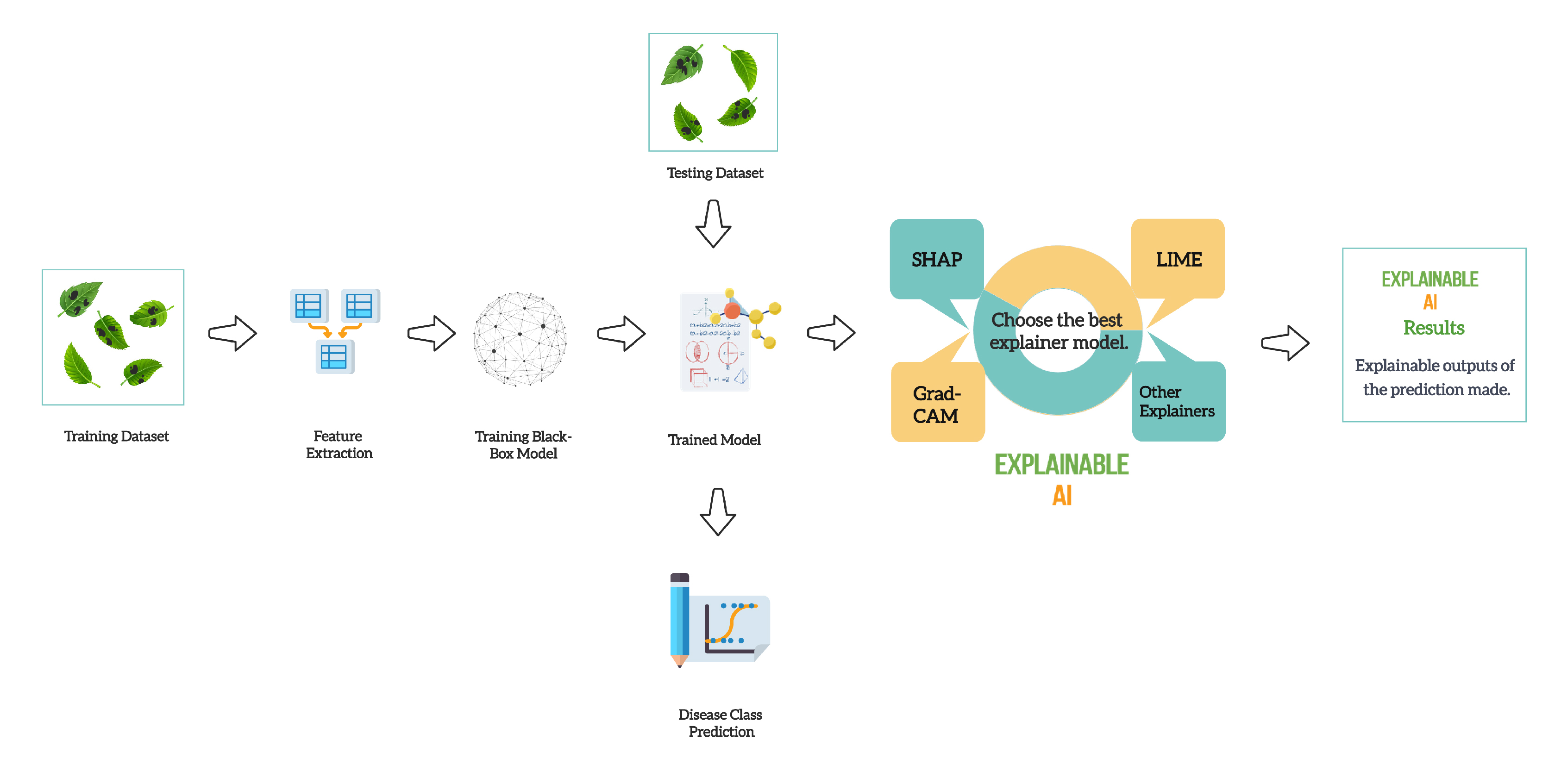}\\
\caption{Depicting XAI solution for leaf-based plant disease detection.}
\label{fig:}
\end{figure*}

To guarantee trust in the system's judgment, Arvind et al. \cite{arvind_deep_2021} developed a Computer-Aided-Diagnostic System (CADS) with an Explainable AI pipeline in 2021. The learning was divided into two parts by the authors. In the first part, deep neural network models are trained with the Transfer learning technique using the Plant Village dataset's original and enhanced data, with 16,684 and 53,476 tomato leaf images, respectively. Among the models used for transfer learning, EffecientNet B5 outperformed VGG16, VGG19, ResNet, Inception v3, MobileNet, and EfficientNet. In the second part, explainable AI methods (LIME and Grad-CAM) were used to interpret the EffecientNet B5 predicted output, and YOLOv4 was used to validate the interpretation. Another Explainable AI-based approach for deep-learning-based disease detection was put forth by Kinger et al. \cite{kinger_explainable_2021} in the same year. They located the disease and highlighted the critical leaf regions for classifying it using the GradCAM++ (Gradient-weighted Class Activation Mapping++) explainer model. They began by utilizing the Plant Village Dataset, which contains images of healthy and infected plant leaves.
Before being used for model training and predictions, the images in the dataset were downscaled to 256x256 pixels. The authors used a pre-trained ImageNet model with a VGG16 deep-learning architecture.

Furthermore, the model's final layers were trained again on the used dataset to improve training accuracy. The authors then employed the visualization method of Grad-CAM++ for model interpretation, which successfully extracted the infected regions and precisely highlighted them with heatmap overlays. Spotty and flaky infected areas on the leaf have been used to validate the results for multiple features of varying sizes in the same image.

The most well-known explainer models, including SHAP, LIME, Grad-CAM, and others, are used to create a visual representation in Fig. 6 that demonstrates how the model will function with the idea of Explainable AI. This proposed generic model will incorporate various explainer techniques to recognize salient areas of the leaf image (infected region) and provide appropriate explanations based on features extracted from the image and predictions made by the black-box model. To provide adequate explanations, various techniques employ various features. For example, LIME acts as an "explainer" to explain "the prediction for a single data sample by altering the feature values and observing the impacted output, which takes the form of local interpretability." Grad-cam is a post-hoc attention method "that creates heatmaps on trained neural-network models and generates maps highlighting regions of interest to support prediction by using feature selection."

\section{\textbf{FUTURE DIRECTIONS}}
We have mentioned some points below that may assist in improving and enhancing the existing state-of-the-art and give some fresh ideas to future scholars.

\subsubsection{\textbf{Disease Stage Identification}}
Each infected plant leaf progresses through four stages, from healthy to diseased (Healthy, Initial, Middle, and Late). Most researchers are only interested in the type of leaf disease, but none have ever focused on the stage of the disease. Diagnosis of infected leaves in the early stages (initial and middle) can assist agricultural practitioners in preventing disease spread. Thus, a system that can recommend specific measures based on disease stages must be developed.

\subsubsection{\textbf{Multiple Disease Infection}}
Most of the time, researchers concentrate on finding just one kind of disease in a plant. On the other hand, a plant can get simultaneously infected by several diseases at once. As a result, a solution to this problem must be discovered so that each disease can be precisely detected and preventive measures can be implemented with caution.

\subsubsection{\textbf{Quantification of a Disease}}
Only a few scientists can determine the level of disease damage on a plant leaf. The proportion of a leaf infected with a given disease will be determined using quantitative methods. This information can be beneficial since remedial steps can be implemented to help with pesticide control. Usually, farmers treat diseases with pesticides without any diagnosis or quantification. As a result, it is necessary to discover a method that will help determine whether a particular pesticide is required.

\section{\textbf{CONCLUSION}}

This paper surveys diverse plant leaf diseases and discusses different leaf disease detection methods and various datasets for plant leaf disease detection. We have also discussed how Machine-Learning and Deep-Learning based (Black-Box) solutions have aided in disease detection in plant leaves. We know that if these diseases are not detected correctly and on time, they can reduce crop yield, eventually leading to long-term issues such as famine. Based on our review of the proposed solutions, we can conclude that there are numerous methods for detecting plant leaf diseases, each with advantages and disadvantages. According to the results of this survey, deep-learning-based solutions such as EffecientNet, DenseNet, and GoogleNet outperform the other proposed solutions. But, as we all know, even these best-performing solutions are the most difficult ones to explain, eventually making the model's predictions untrustworthy. Black-box models are inappropriate for specific applications, such as in the medical domain, where incorrect system decisions can be highly harmful, similar to our problem. Therefore, explainability was presented as a requirement for resolving legal issues due to the increased use of AI systems. To address this problem, we propose Explainable AI (XAI), which focuses on making AI solutions transparent to be trusted and presented to users more understandably.

We seek to integrate deep-learning-based approaches with the visual-explainable techniques offered by XAI in this latest development in leaf disease detection to find a suitable explanation for the judgments made by the black-box model. Based on the disease detected by the black-box model, visually explainable approaches like LIME, GradCAM, and GradCAM++ have performed remarkably well in accurately localizing the infected regions of the leaf image and highlighting them with heatmap overlays. Although the findings of the XAI method used to detect plant diseases are encouraging, we know the field of XAI is still developing, and the explanation maps are not enough to support decision-making. 

Furthermore, future studies will concentrate on making such interpretations logically correct and verbose because visual interpretations, in particular, are inherently biased by humans.

\section{Competing Interests}
Authors declare no conflict of interest.

\section{Funding Information}
Saurav Sagar recieved partial support from Ministry of Education (GATE Fellowship), Govt. of India.

\section{Author contribution}
All authors whose names appear on the submission

\begin{itemize}

\item made substantial contributions to the conception or design of the work; or the acquisition, analysis, or interpretation of data; or the creation of new software used in the work;

\item drafted the work or revised it critically for important intellectual content;

\item approved the version to be published; and

\item agree to be accountable for all aspects of the work in ensuring that questions related to the accuracy or integrity of any part of the work are appropriately investigated and resolved. 

\end{itemize}

\section{Data Availability Statement}
Data sharing not applicable to this article as no datasets were generated or analysed during the current study.

\printbibliography
\end{document}